# A Novel Solution of an Elastic Net Regularization for Dementia Knowledge Discovery using Deep Learning

Kshitiz Shrestha[1], Omar Hisham Alsadoon[2], Abeer Alsadoon[1, 3,4,5], Tarik A. Rashid[6], Rasha S. Ali[7], P.W.C. Prasad[1,4], Oday D. Jerew[5]

[1]School of Computing and Mathematics, Charles Sturt University (CSU), Wagga Wagga, Australia
[2]Department of Islamic Sciences, Al Iraqia University, Baghdad, Iraq
[3]School of Computer Data and Mathematical Sciences, University of Western Sydney (UWS), Australia
[4]Kent Institute Australia, Information Technology Department, Sydney, Australia
[5]Asia Pacific International College (APIC), Information Technology Department, Sydney, Australia
6Computer Science and Engineering, University of Kurdistan Hewler, Erbil, KRG, Iraq
[7]Department of Computer Techniques Engineering, AL Nisour University College, Baghdad, Iraq

**Abeer Alsadoon[1*]**

[1]School of Computing and Mathematics, Charles Sturt University, Sydney, Australia
* Corresponding author. Dr. Abeer Alsadoon, Charles Sturt University, Sydney Campus, Sydney, Australia, Email: alsadoon.abeer@gmail.com , Phone +61 2 9291 9387

## Abstract

*Background and Aim:* Accurate classification of Magnetic Resonance Images (MRI) is essential to accurately predict Mild Cognitive Impairment (MCI) to Alzheimer's Disease (AD) conversion. Meanwhile, deep learning has been successfully implemented to classify and predict dementia disease. However, the accuracy of MRI image classification is low. This paper aims to increase the accuracy and reduce the processing time of classification through Deep Learning Architecture by using Elastic Net Regularization in Feature Selection. *Methodology:* The proposed system consists of Convolutional Neural Network (CNN) to enhance the accuracy of classification and prediction by using Elastic Net Regularization. Initially, the MRI images are fed into CNN for features extraction through convolutional layers alternate with pooling layers, and then through a fully connected layer. After that, the features extracted are subjected to Principle Component Analysis (PCA) and Elastic Net Regularization for feature selection. Finally, the selected features are used as an input to Extreme Machine Learning (EML) for the classification of MRI images. *Results:* The result shows that the accuracy of the proposed solution is better than the current system. In addition to that, the proposed method has improved the classification accuracy by 5% on average and reduced the processing time by 30 ~ 40 seconds on average. *Conclusion:* The proposed system is focused on improving the accuracy and processing time of MCI converters/non-converters classification. It consists of features extraction, feature selection, and classification using CNN, FreeSurfer, PCA, Elastic Net, Extreme Machine Learning. Finally, this study enhances the accuracy and the processing time by using Elastic Net Regularization, which provides important selected features for classification.

# 1. Introduction

Alzheimers disease(AD) is one of the most common causes of dementia in today's world. Usually, the symptoms of Alzheimer's are visible after 60 years of age.However, some forms of AD develop very early (30-50 years) for individuals having gene mutation. Alzheimer's disease gives rise to structural and functional changes in the brain (M. TANVEER et al, 2019). In over 60% of dementia cases, Alzheimer disease is a neurodegenerative disease that causes progressive and disabling impairment of cognitive functions including memory, comprehension, language, attention, reasoning, and judgment (Yi Tang et al, 2018).It can also cause the death of the person. The cause of AD is still undefined. So, the early detection of AD is very crucial to apply protective measures to slow down the development of AD. In the past, functional neuroimaging techniques are used to classify AD patients. Functional imaging gives the images as patients complete the task. The areas of the brain that participate in completing the task give the researchers the visual 3-D view of the parts of the brain. However, the traditional method has low accuracy because these images have an invasion, low resolution, and not much effective means for diagnosis of brain diseases. In this study, structural imaging is used to provide images of the brain's anatomical structure. It is used primarily because of the high resolution, non-invasion, and very useful for the diagnosis of brain disease like dementia (L. Weiming et al, 2018). The latest technology CNN and Free Surfer are survived as features extraction methods from MRI images. Elastic Net and PCA are used for feature selection and Extreme Machine Learning for classification purposes (M. Liu et al, 2018).

The machine learning algorithm performs impressively well in the classification and prediction of MCI to AD conversion. Deep learning is the latest and promising machine learning methodology used for classifying and identifying the patterns of images (L. Weiming et al, 2018). CNN is the deep learning's most widely used architecture which provides advantages for the classification and prediction of dementia disease. CNN consists of several layers such as the pooling layer, convolutional layer, and a fully connected layer. CNN is used for learning and extracting the features of the MRI image dataset and has been implemented successfully in the area of diagnosis of AD (A. Chaddad et al, 2018). Therefore, MCI to AD conversion's prediction can be achieved using CNN. However, CNN is still sensitive in terms of the number of datasets, regularization, filter size, activation function, and the number of feature maps to classify MRI images. Thus, the current technology still has some limitations which result in low accuracy, weight matrices, and high processing time of the neural network. The model can be defined as the best model if it improves the accuracy with less processing time and minimizes the weight matrices of the neural network (M. Liu et al, 2018) Current studies of CNN based features learning and extraction uses various algorithm and techniques for improvement of the system's accuracy. The result of the state-of-art method produces the classification accuracy of 81.4% and Area Under Curve (AUC) of 87.7% (L. Weiming et al, 2018). The classification accuracy of CNN is 8.06% higher than that of the Mini-Mental Status Examination alone (W. Huang et al, 2018). The LASSO regression used in the state-of-art method causes losing of relevant independent variables and inconsistency in feature selection. Therefore, current studies still have an area that needs to be improved.

The purpose of this paper is to enhance the accuracy and processing time of the system by combining Lasso regression and ridge regression. This research aims to increase the performance of the system by applying regularization that prevents the overfitting of training datasets. This study proposes an Elastic Net Regularization is the combination of Lasso regression and modified ridge regression. It forms a group if there is a correlated variable to protect the valuable information from being removed. It maintains the consistency in feature selection by including all the relevant independent variables. Thus, it improves the accuracy and the processing time of the prediction model of MCI to AD conversion.

Accurate classification of Magnetic Resonance Images (MRI) is an important thing to predict Mild Cognitive Impairment (MCI) accurately in Alzheimer's Disease (AD) conversion. At the same time, deep learning has been successfully carried out in classifying and predicting dementia disease. However, the accuracy of MRI image classification is low. So that, the main goal of this research represented in increasing accuracy and reducing the processing time of classification via Deep Learning Architecture by the usage of Elastic Net Regularization in Feature Selection. The Elastic Net Regularization adds a quadratic component to the penalty to consist of all the relevant independent variables. It creates a set if there is a correlated variable to protect the valuable information from being removed.

The remainder of the paper is organisedas follows. Section 2 discusses the literature review techniques. The proposed method is presented in Section 3 with details of the proposed equation, area of improvement, and Why Elastic Net Regularization. Section 4 discusses the results discussion and conclusion and future work illustrated in Section 5.
.

## 2. Literature Review

The primary purpose of the Literature Review is to do the survey of present papers and find new ways to improve the current systems. It provides a perception of different methods, techniques, tools that have been applied in the area of this study. In addition to that, this section provides a review of different papers and research on the related perspective area.

(J. Ding et al, 2018) Enhanced the accurate detection of dementia degree in the brain. This research provides deep integration of visual information system solution that solves the main two medical image quality problems. Then, the dementia image signals have been represented in matrixes, and the brain signals of dementia induction are selected and reconstructed. This study has improved the precision of detecting the dementia degree in brain diseases. The solution has used the two algorithms using java for deep visual perception for improving the quality of an image. The proposed solution provides an analysis of might affect the infrequent signals of the brain image and removes them to obtain an adaptive dynamic dementia image signal. (A. Chaddad et al, 2018) analyzed the radiomic features obtained from individual subcortical brain regions to identify Alzheimer's Disease. The authors use the extracted 45 radiomic features and produced 21 feature maps using Convolutional Neural Network. Wilcoxon test and Random Classifier were used to identify dominant regions which are correlation and volume in the hippocampus and amygdala. The proposed framework provides an Area Under Curve of 92.58% using entropy features. However, it applies only 235 number of a dataset from the Open Access Series of Imaging Studies (OASIS) database and uses only MR images. As a feature work to this study, we can use different ways for the encoding of the texture by utilizing Convolution Neural Networks which can improve the classification accuracy. (H. Choi et al, 2018) combined the advantages of the Mini-Mental Status Examination and Korean Longitudinal Study on Cognitive Aging and Dementia. This study offers a solution using 435 dementia patients and 2666 cognitively normal elderly individuals. The K-Nearest Neighbor has been used for information recovery and developed a deep neural network for classification. The proposed solution provides an 8.06% higher accuracy than using just the Mini-Mental Status Examination alone. However, this research has been used only binary classification. For that, we can use a multi-level classification. For future work, the model needs to identify the meaning of the hidden patterns through Artificial Intelligence. [7] developed a Deep Convolutional Neural Network which uses Alzheimer's Disease and Normal Control subjects for training. This research provides a solution using Mild Cognitive Impairment subject as input for prediction purposes which predicts the Mild Cognitive Impairment to Alzheimer's Disease conversion. Besides, it provides accuracy, sensitivity, and specificity of the Convolutional Neural Network for the classification of Mild Cognitive Impairment conversion as 81.0%, 87.0%, and 84.2%, respectively. It provides 84.2% accuracy in Mild Cognitive Impairment conversion producing a better result than the Support Vector Machine and Volume of Interest. In contrast, it uses minimally processed images without spatial normalization which might effect the accuracy, so that needs to be considered to increase the accuracy through using spatial normalization. As future work, the proposed system can be applied for Positron Emission Tomography images and use a more significant number of datasets.

(J. Liu et al, 2016) developed the artificial neural network that uses multi-view fusion that combines with multiple estimators. This research paper provides a two-stage binary classification, i.e., Alzheimer's Disease and Normal Control and Mild Cognitive Impairment and Normal Control. Ensemble learning layer and the soft-max layer is present in it to predict either the given subject is Alzheimer's Disease or Mild Cognitive Impairment. Among the three views Magnetic Resonance Imaging + Positron Emission Tomography gives the best result. However, the proposed solution only focuses on binary classification, so it is necessary to extend the framework to the multi-class classification problem. (W. J. Niessen, 2016) studied the strong influence of genetics in the brain for knowledge discovery from Magnetic Resonance Imaging images. The author uses the extracted data by data references, the aging model, and computer-aided for the diagnosis. The proposed solution makes analyzes to the brain Magnetic Resonance Imaging images in various aspects of a brain using the methodology like Imaging genetics, Ageing brain models, quantitative imaging biomarker, and Machine learning. However, the proposed solution has not done any measurement which should be considered to get the accuracy for comparison purposes.

(C. Ieracitano et al, 2018) developed a customized Convolutional Neural Network using Rectified Linear Unit for getting better accuracy. The authors use a 2-Dimensional Power Spectral density images as an input to a customized Convolutional Neural Network that performs two ways or/and three ways of classification. The proposed framework provides an accuracy of 89.8% in two ways of classification and 83.3% in three ways classification. In contrast, there is a limitation of non-stationarity of the electroencephalogram epochs of each patient under analysis. This issue will result in misclassification and will effect overall the accuracy of the proposed system. As proposed future work, the system needs a Power Spectral density-based Convolutional Neural Network architecture that will be trained using more powerful graphics processing units as well as, a large number of medical images is needed for the processing stage. (R. Ju et al, 2017) developed a deep learning algorithm that has improved the prediction accuracy and minimizes the standard deviation. This study solves the problem by using brain networks and clinically related information for the detection of Alzheimer's Disease in the initial stage. The proposed research provides an improvement of about 31.21% in prediction accuracy and minimizes the standard deviation by 51.23%. As future work, the proposed system needs to be applied to more massive datasets and use it for the diagnosis of other neurological diseases.

(W. Huang et al, 2018) enhanced a system to detect dementia disease using low-resource pairwise learning. The authors use a deep learning algorithm on Arterial Spin Labelling (ASL) magnetic resonance images. As future work, the model needs to work more on medical imaging problems using sophisticated deep learning techniques. (J. E. Lee et al, 2016) examines Nucleus basalis magno-cellularis stimulation to study the changes in spatial memory and neurotransmitter systems. The authors use 192 IgG-saporin for basal forebrain cholinergic neuron degeneration associated with memory and learning. The proposed system provides a piece of new information that consolidation and retrieves the visuospatial memory which is enhanced by Nucleus basalis magnocellularis stimulation. However, it is not clear how much the electrical stimulation effects and make changes in the neurotransmitter system, and they are related to visuospatial memory. The study of the electrical stimulation mechanism and the neurotransmitter systems will be considered as the future work of this paper. (H. Akinori et al, 2018) constructed a Multiple Layer Perceptron as a predictive model that predicts the outcome of the Morris Water Maze. The authors use the output data of the Morris Water Maze operation and fed it to the Artificial Neural Networks as an input. The output results of the Artificial Neural Network model and Morris Water Maze operation were compared. The prediction accuracy between the human tester and the Artificial Neural Network model is similar. Apart from that, the accuracy of the predictive model is affected when the explanatory and the objective variables are on the same data representation.

(T. Zhou et al, 2018) developed a stage-wise deep neural network for learning features and fusion of multimodality data for prediction of Dementia. This study uses multimodality neuroimaging and genetic data that alleviates the heterogeneity issue to learning the different modality representation using separate deep neural network models. The proposed framework allows the maximum use of training of the available samples of each stage. Moreover, it provides an accuracy of 64.4% for three-class classification, which is enhanced the results by 18% when compared with the current state-of-art methods. However, this research paper focuses only on the Region of interest (ROI) features which limit the richness of structural and functional brain information. For future work, the proposed model needs to use the original imaging data with advanced deep neural networks to utilize its full power. It also needs to incorporate other confounding factors such as gender, education level, etc. to improve the performance of the model. (L. d. Langavant et al, 2018) predicted dementia in population-based surveys without the involvement of the clinical diagnosis of dementia. The proposed solution uses Agglomerative Hierarchical Clustering as an unsupervised machine learning technique to classify the dataset from the Health and Retirement Study. It provided an accuracy of 93.3% and predicted the probability of dementia >.95. But, still, it is difficult to understand from the clinical point of view, and it is like a black box where the actual mechanism remains opaque.

(M. Liu et al, 2018) developed a Deep Multi-task Multi-channel Learning ($DM^2L$) framework for simultaneous Alzheimer's disease classification and regression. The authors offer a solution using extracted multiple image patches from identified landmarks of Magnetic Resonance images as an input along with demographic information incorporated explicitly into a learning process. The proposed solution performs better than other state-of-art solutions in disease classification and clinical score regression. Despite that, the framework trains a model on Alzheimer's Disease Neuroimaging Initiative-1 and tests it on two independent datasets that degrade the performance due to the difference in the data distribution. As a future work to this research, the model needs to automatically learn weights for classification and clinical score regression of disease which could have a different contribution. (L. Weiming et al, 2018) developed a framework that analyses Magnetic Resonance Imaging data only for Mild Cognitive Impairment to Alzheimer's Disease conversion's prediction.

This research provides a solution through using Convolutional Neural Network and FreeSurfer as feature extraction methods. The features extracted from the data will be fed them to the extreme machine learning to classify the converters and non-converters Mild Cognitive Impairment. The proposed system provides an accuracy of 81.4% and Area under the curve of 87.7% which performs better than the state-of-the-art methods. Even though, it gives less specificity when compared to the state–of–the–art method. As future work, the model needs to investigate the combination of other modality data with the proposed method for improving performance.

## 2.1 State of the Art

The highlighted inside broken blue line represents the features of the current system (in fig. 1) and the broken red line highlighted inside represents limitations (in fig. 1). (L. Weiming et al, 2018) developed a Convolutional Neural Network (CNN) based Magnetic Resonance Imaging (MRI) image analysis for Mild Cognitive Impairment to Alzheimer's Disease conversion prediction. They used Convolutional neural network and FreeSurfer as a feature extraction methods, and the extracted features will be proceeded by extreme machine learning to classify the converters and non-converters Mild Cognitive Impairment (MCI) (L. Weiming et al, 2018). The proposed prototype provides an accuracy of 81.4% and Area under the curve of 87.7% respectively. The proposed model consists of four stages (Fig. 1: State of the Art), i.e. the pre-processing stage, features extraction stage, feature selection stage, and classification stage.

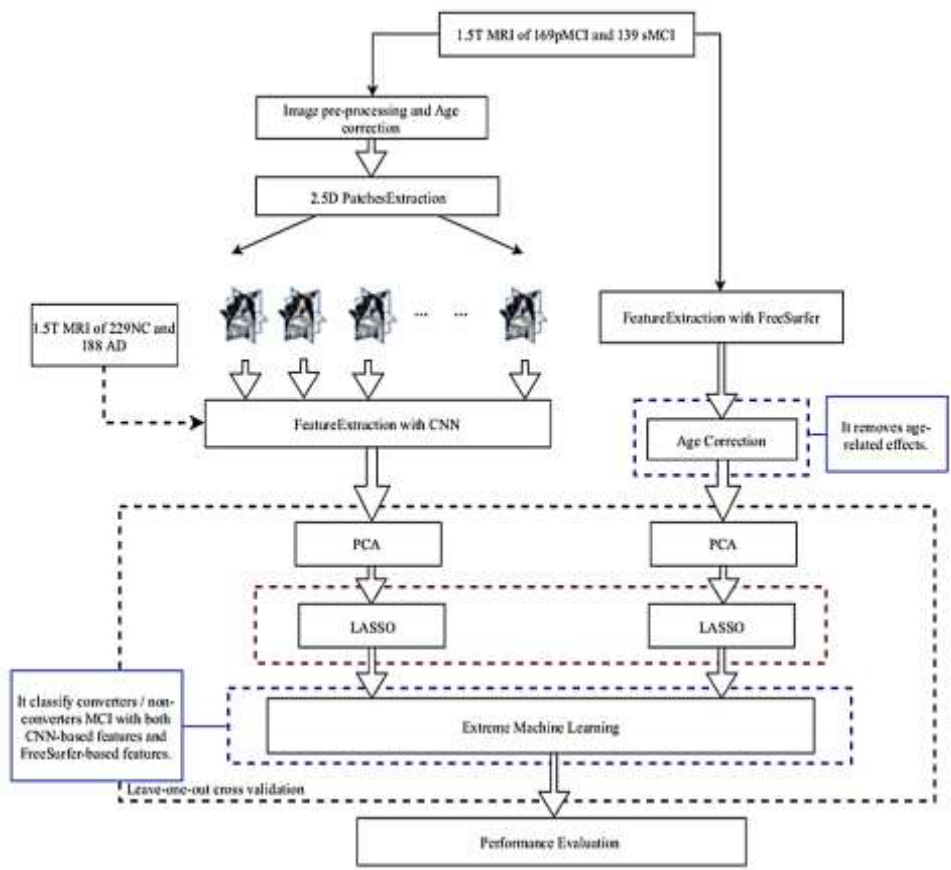

Fig. 1: Block Diagram of the State of the art System, *(L. Weiming et al, 2018)*
[The blue borders show the excellent features of the current state of the art solution, and the red border refers to the limitation of it]

*Pre-processing Stage:* Magnetic Resonance Imaging dataset is taken from Alzheimer's Disease Neuroimaging Initiative-1. These images are skull-stripped and then a B-spline free-form deformation registration used for aligning to the MNI151 templates. Images are registered using Tong's way (L. Weiming et al, 2018) to classify AD/NC and converters/non-converters that have a control point spacing between 10 and 5 mm after the effect of deformable registration. The image intensity of the subjects was normalized to match the MNI151 templates. A linear regression model is fitted to estimate Age correction as shown in Fig. 1.

*Features Extraction Stage:* In this stage, CNN and Freesurfer have utilized as a feature extraction methods from the MRI images. Three patches of 32 x 32 pixels from a transverse, coronal, and sagittal plane which centered at the same point are extracted to form a 2.5D patch in MRI. Then a 2-D RBG patch is obtained after combining these three patches. CNN has three pooling layers and three convolutional layers (L. Weiming et al, 2018) which takes 32 x 32 RGB patch as an input. Total 32 feature maps with a size of 32 x 32 are generated using the first convolutional layer which is down-sampled into 16 x 16. Finally, 64 features maps with a size of 4 x 4 are generated by the remaining two layers. A feature vector is produced after concatenation of features, and then it is fed into a fully connected layer and soft-max layer for classification. For each image 154624 (1024 x 151) features are generated by CNN. Additionally, other features have been obtained from the MRI images by using FreeSurfer such as cortical volume, surface area, cortical thickness average, and standard deviation of the thickness in each region of interest (L. Weiming et al, 2018).

*Feature Selection Stage:* The principal component analysis (PCA) and the Least Absolute Shrinkage and Selection Operator (LASSO) (L. Weiming et al, 2018) were used for feature selection to minimize the total number of features. In this model, there is no satisfactory result has been achieved in accuracy and processing time due to the fact of using the LASSO method in the Feature Selection stage for minimizing the number of features. The use of the LASSO method causes losing some relevant independent features and inconsistency in feature selection due to the selection of only one feature from all correlated features. Consistent feature selection and the inclusion of the relevant independent features are essential for getting high accuracy and low processing time. If the features that play a vital role in the prediction model are removed in feature selection then it will be difficult to get the high accuracy of the model. Even though we get a reasonable value in the accuracy of the prediction model, the desired value is not achieved. As a result, the system is not fully optimized, and lower values are obtained due to the lack of relevant features.

*Classification Stage:* With the help of Extreme Machine Learning (L. Weiming et al, 2018), outputs are calculated so that random generation of input weight matrix is avoided and classifies converters/non-converters MCI with both CNN-based features and FreeSurfer-based features.

This model provides an accuracy of 81.4% and processing time of 359ms respectively (L. Weiming et al, 2018). The LASSO regression is implemented in Feature Selection Stage before proceeding the features to Extreme Machine Learning to reduce the final number of features as shown in equation 1 (L. Weiming et al, 2018). However, still, accuracy and processing time can be enhanced by using the other features selection methods.

$$L(W) = \min_{\propto} 0.5||y - D \propto ||_2^2 + \lambda|| \propto ||^1 \tag{1}$$

Where,
$y \in R^{1 \times N}$ is the vector consisting of *N* labels of training samples,
$D \in R^{N \times M}$ is the feature matrix of *N* training samples consisting of *M* features,
$\lambda$, is the penalty coefficient
$\propto \in R^{1 \times M}$, is the sparse target coefficient

**Table 1: LASSO Regression**

| |
|---|
| **Algorithm:** LASSO regression to minimize the total number of features |
| **Input:** a Large number of features |
| **Output:** Less number of features |
| 1: **BEGIN** |
| 2: **Create function** Lasso (ipy, ipx, λ, N) |
|      # ipy -> Inner product vector, ipy$_i$ = <y, X$_{\cdot i}$> |
|      # ipx -> Inner product matrix, ipx$_{ij}$ = <X$_{\cdot i}$, X$_{\cdot j}$> |
|      # λ -> Penalty coefficient |
|      # N -> Number of samples |
| 3: **define** threshold variable stop_thr      # Threshold for stopping iteration |
| 4: **define** variable length (ipy) p |
| 5: **define** variable beta = 0 with length p |
| 6: **define** variable gc = 0 with length p      #Gradient component vector |
| 7: **start do-while loop from here** |
| 8:    **assign** variable difBeta$_{max}$ = 0 |
| 9:    **start for loop** j = 1 to p |
| 10:      **assign** variable z = (ipy [j] – gc [j])/N + beta [j] |
| 11:      **assign** beta_tmp = max (0, z- λ) – max (0, -z- λ) |
| 12:      **assign** difBeta = beta_tmp – beta [j] |
| 13:      **assign** difabs = abs (difBeta) |
| 14:      **apply** i**f then condition with** difabs>0 |
| 15:        if true then, **assign** beta [j] = beta_tmp and |
| 16:        **assign** gc = gc +ipx[j] x difBeta and      #Update gradient components |
| 17:        **assign** difBeta$_{max}$ = max (difBeta$_{max}$, difabs) |
| 18:      **end if then condition** |
| 19:    **end for loop** |
| 20: **putting while** condition (difBeta$_{max}$ >= stop_thr) in do-while loop |
| 21: **end do-while loop** |
| 22: **return the value beta**9 |
| 23: **end function** |
| 24: **END** |

**Flowchart of LASSO regression:**

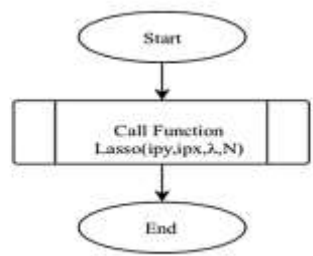

Fig. 2: Flowchart of Function Lasso

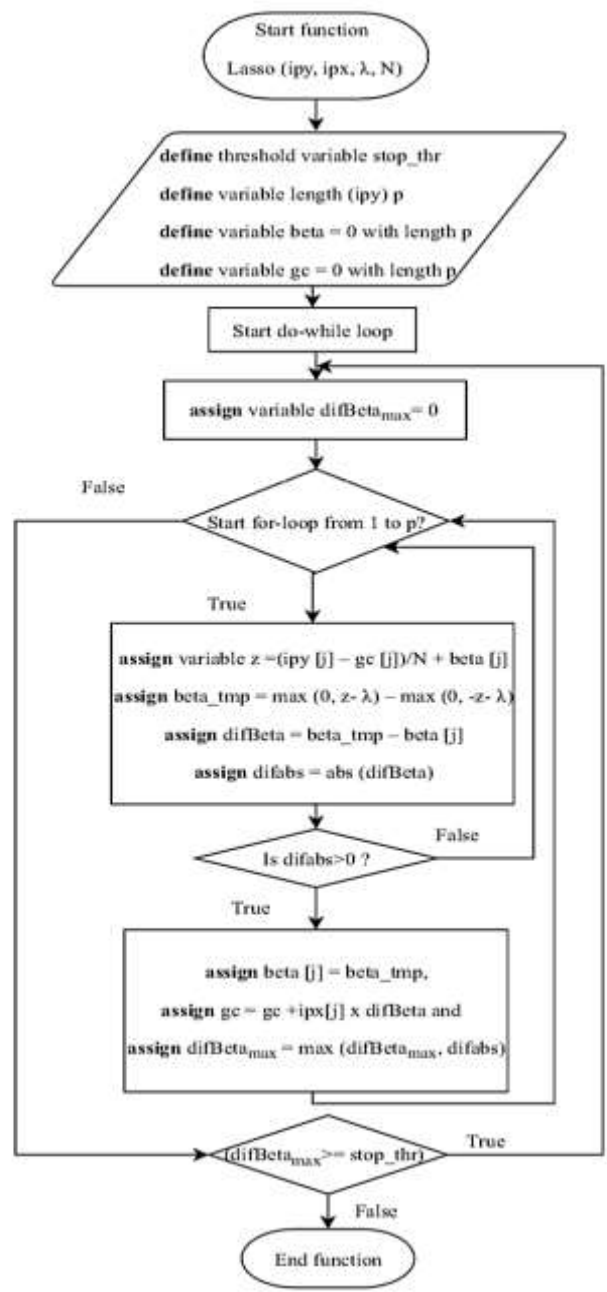

Fig. 3: Flowchart of Lasso regression

## 3. Proposed System

After reviewing and testing several features extraction and classification techniques by using various deep learning algorithms, we analyzed the pros and cons of each method. Based on the obtained results, the accuracy and the processing time are the key factors to be considered in this research. From the reviewed list of features extraction and classification techniques, we select the best one, (L. Weiming et al, 2018) as the basis for our proposed solution. The main reason behind choosing this research work is that (L. Weiming et al, 2018) has used Extreme Machine Learning for classification which classifies converters/non-converters with both CNN-based features and FreeSurfer-based features. The proposed classification technique adopts kernel to calculates the outputs and avoids the random generation of the input weight matrix. In addition to that, the classification accuracy and area under a curve of the (L. Weiming et al, 2018) model gives higher results when compared to the traditional methods like SVM, Random Forest. The next reason for using the model of [4] is the use of the Age Correction method by fitting the linear regression model to remove age-related effects. As a result, the using of Age Correction method has improved the accuracy and area under a curve of the model. On the other hand, the proposed model of (L.

Weiming et al, 2018) has adopted the LASSO regression method as a feature selection method. By analyzing the results, the LASSO method causes losing some relevant independent features and inconsistency in feature selection due to the selection of only one feature from all correlated features. Furthermore, various other research works were reviewed and analyzed related to feature selection technique. The solution proposed by (M. Liu et al, 2018) has provided a method to overcome the limitation of LASSO regression for feature selection. This paper enhances the accuracy and processing time of Converters/Non-Converters MCI classification. The proposed system consists of four major stages (Fig. 4) called pre-processing stage, feature extraction stage, feature selection, and a classification stage.

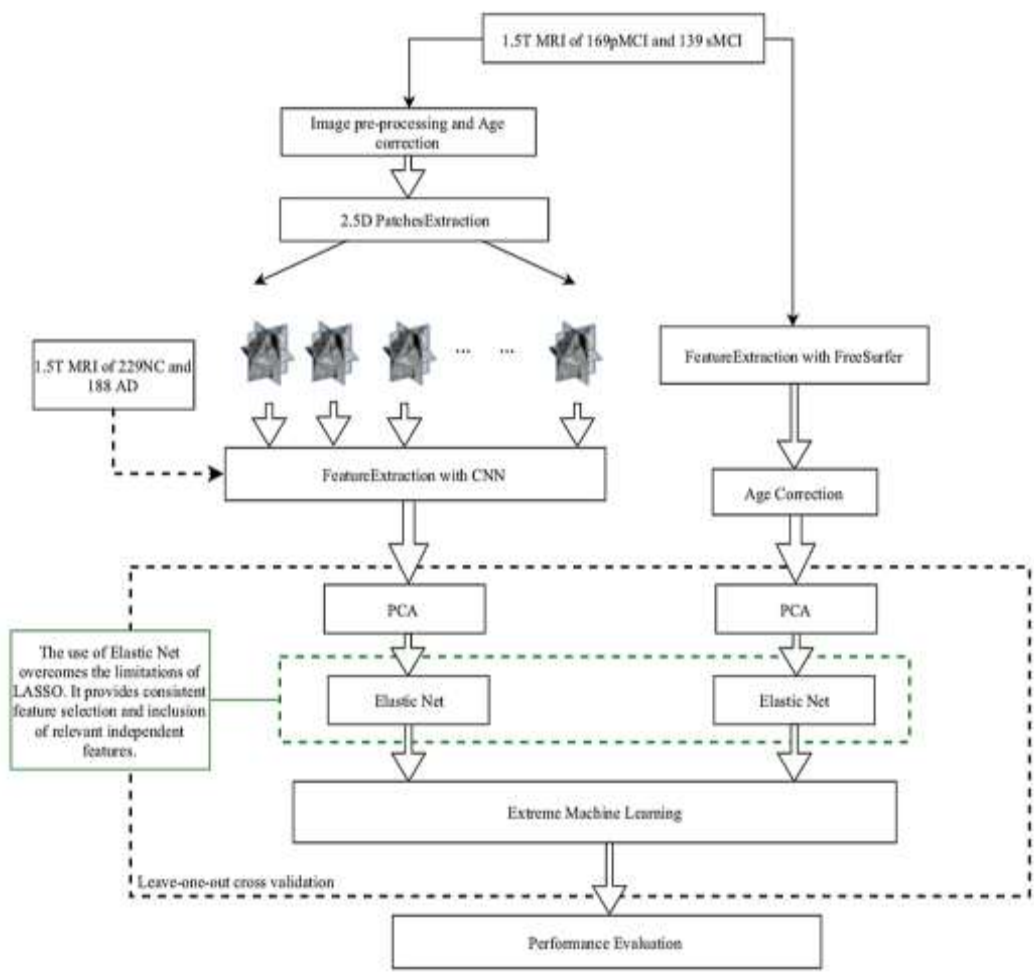

Fig. 4: Block diagram of the proposed system for the classification of MCI using Elastic Net Regularization
[The green borders refer to the new parts in our proposed system]

*Pre-processing Stage:* Magnetic Resonance Imaging dataset is taken from Alzheimer's Disease Neuroimaging Initiative-1. These images are skull-stripped and then a B-spline free-form deformation registration used for aligning to the MNI151 templates. Images are registered using Tong's way (L. Weiming et al, 2018) to classify AD/NC and converters/non-converters that have a control point spacing between 10 and 5 mm after the effect of deformable registration. The image intensity of the subjects was normalized to match the MNI151 templates. A linear regression model is fitted to estimate Age correction as shown in Fig. 1.

*Features Extraction Stage:* In the features extraction stage, the CNN and the Freesurfer are used for features extraction process from MRI images. Three 32 x 32 patches from transverse, coronal, and sagittal plane centered at the same point are extracted to form a 2.5D patch in MRI. Then a 2D RBG spot is obtained after combining these three patches. CNN has three pooling layers and three convolutional layers (L. Weiming et al, 2018) which takes 32 x 32 RGB patch as an input. Total 32 feature maps with a size of 32 x 32 are generated by using the first convolutional layer which is down-sampled into 16 x 16. Finally, 64 features maps with a size of 4 x 4 are generated by the remaining of the two layers. A feature vector is produced after concatenation of features, and then it is fed into a fully connected layer and soft-max layer for classification. For each image 154624 (1024 x 151) features are generated by CNN. Besides, more morphological information of MRI images is mined by using

FreeSurfer such as cortical volume, surface area, cortical thickness average, and standard deviation of the thickness in each region of interest (L. Weiming et al, 2018).

Feature Selection Stage: The principal component analysis (PCA) and Elastic Net Regularization are used for feature selection to minimize the total number of features. Elastic Net Regularization provides consistent feature selection and inclusion of relevant independent features and removing the limitations of the LASSO feature selection method.

*Classification Stage:* With the help of Extreme Machine Learning (L. Weiming et al, 2018), outputs are calculated so that random generation of input weight matrix is avoided and classifies converters/non-converters MCI with both CNN-based features and FreeSurfer-based features.

The Elastic Net Regularization is used in the proposed system for a feature selection process. The main purpose of using Elastic Net Regularization is that it solves the limitations of Lasso regression in losing the relevant and independent features. It adds a quadratic part to the penalty which is obtained after modifying the Ridge Regression. The proposed method provides consistent feature selection and inclusion of relevant independent features which are important for getting high accuracy and low processing time.

The derivation for a new equation is proposed. The new equation has a quadratic part of the Ridge regression that is added to the Lasso regression. The achieved results in forming a group of the correlated variable instead of selecting only one variable and maintains the consistency in feature selection by including all the relevant independent variables. The proposed equation of Elastic Net Regularization is the combination of both Lasso Regression and modified Ridge Regression and expressed as:

$$EL(W) = L(W) + MR(W) \quad (2)$$

Where,
EL(W) is the Elastic Net Regularization,
L(W) is the Lasso Regression and
MR(W) is the Modified Ridge Regression

Ridge regression has been adopted from the second-best solution to get an important part of its equation. The chosen part from the second-best solution in the state of the art is combined with the LASSO Regression to get the proposed mathematical equation. The mathematical formula of Ridge Regression (L2 Regularization) (M. Liu et al, 2018) is as follows:

$$R(W) = \min_{\propto} 0.5||y - D \propto ||_2^2 + \lambda || \propto ||_2^2 \quad (3)$$

Where,
$y \in R^{1 \times N}$ is the vector consisting of $N$ labels of training samples,
$D \in R^{N \times M}$ is the feature matrix of $N$ training samples consisting of $M$ features,
$\lambda$, is the penalty coefficient
$\propto \in R^{1 \times M}$ is the sparse target coefficient

As it is described in the section above, the Modified Ridge Regression is obtained from Ridge Regression after taking only the precise part of the equation and involved it with the LASSO Regression to form the proposed solution and solve the problem. The modified Ridge Regression is expressed as follows:

$$MR(W) = \lambda || \propto ||_2^2 \quad (4)$$

Where,
$\lambda$, is the penalty coefficient
$\propto \in R^{1 \times M}$, is the sparse target coefficient

## 3.1 AREA OF IMPROVEMENT

In this research, a new mathematical equation was proposed that could help the feature selection method to reduce the number of features with improving the system accuracy and processing time by providing relevant and independent features. However, the state of the art model shows a low level of accuracy due to the lack of relevant features. On the other hand, the use of Elastic Net Regularization adds a quadratic part to the penalty which when used it alone is called the Ridge regression as it was discussed in the previous section. The major limitation of state of the art (L. Weiming et al, 2018) represented in the loss for some relevant independent features and inconsistency in feature selection as it selects only one feature from all correlated features. The proposed solution is a combination of Lasso regression and Ridge regression. The Lasso and Modified Ridge regression are included in Elastic-net regularization. It forms a group if there is a correlated variable to protect the valuable information from being removed. As well as, it maintains the consistency in feature selection by including all the relevant independent variables. The accuracy of the system is improved as it solves the problem of over-fitting. **Increasing the accuracy and reducing the processing time is crucial to get better performance. By using the proposed system, the extracted important features are included and used for classification purposes, and this helps to improve the performance of the system. Thus, such features enhance the information available for the classification, increasing the accuracy and reducing the processing time of the system.** . Based on the literature review of this paper, the available solutions that classify the MCI converters/ non-converters have used Lasso regression. None of these solutions has considered the correlated variables and relevant independent variables. Our proposed solution has formed a group of correlated features by using modified Ridge Regression. It increases the system accuracy by including all the relevant independent variables. See Table 2 and Fig. 4.

**Flowchart of Elastic Net**

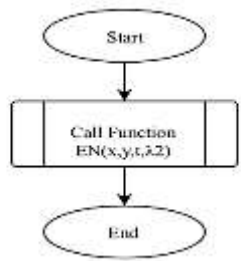

Fig. 5: Flowchart of Function Elastic Net

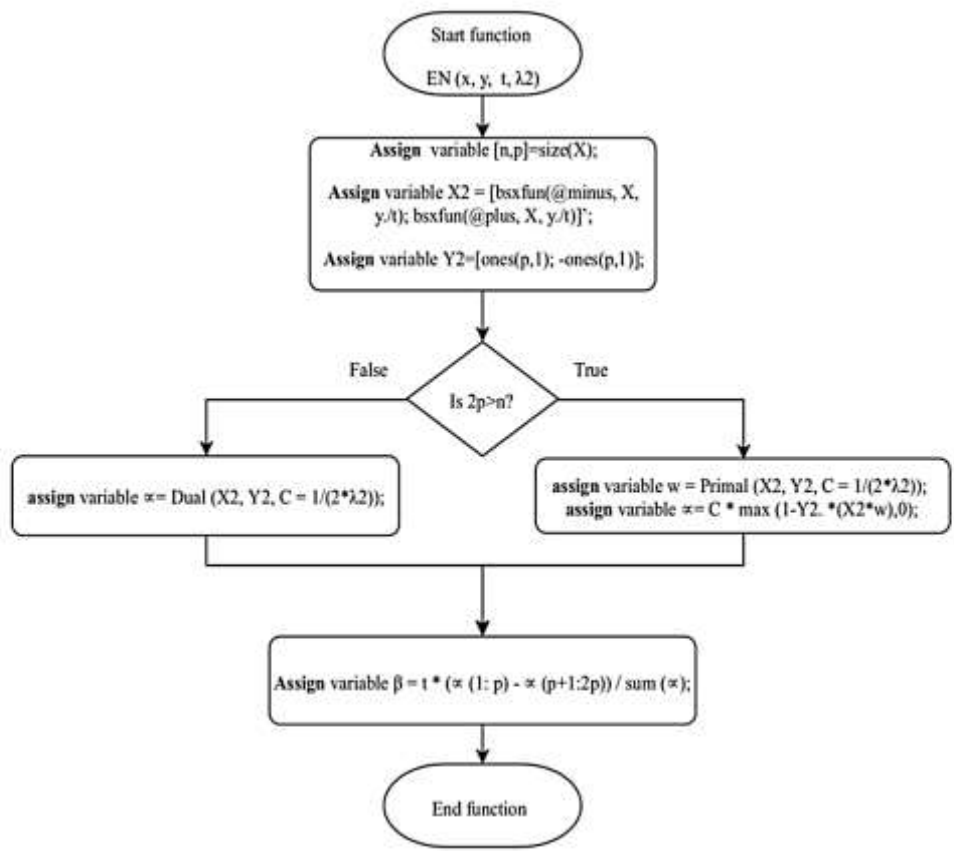

Fig. 6: Flowchart of Elastic Net Regularization

**Table 2: Proposed Elastic-net regularization for MCI classification**

| |
|---|
| **Algorithm:** Proposed Elastic-net regularization |
| **Input:** A Large number of features |
| **Output:** Less number of features |
| 1: **BEGIN** <br> 2: **Create Function** EN (x, y, t,λ2); <br>   # x -> Inner product matrix <br>   # y -> Inner product vector <br>   # t -> Number of samples <br>   # λ2 -> Penalty coefficient <br><br> 3: **Assign** variable [n,p]=size(X); <br> 4: **Assign** variable X2 = [bsxfun(@minus, X, y./t); bsxfun(@plus, X, y./t)]'; <br><br> 5: **Assign** variable Y2=[ones(p,1); -ones(p,1)]; <br> 6: **Apply if then else condition** with (2p>n) <br> 7:    **if true then**, **assign** variable w = Primal (X2, Y2, C = 1/(2*λ2)); and <br> 8:    **assign** variable ∝ = C * max (1-Y2. *(X2*w),0);   # Target sparse coefficients <br> 9: **else** <br> 10: **if false then**, assign variable ∝ = Dual (X2, Y2, C = 1/(2*λ2));   # Target sparse coefficients <br> 11: **end if then else** condition <br> 12: **Assign** variable β = t * (∝ (1: p) - ∝ (p+1:2p)) / sum (∝); <br> 13: **return** the value β <br> 14: **end function** <br> 15: **END** |

# 4. Results and Discussion

Python 2.7.10 with TensorFlow were used for the implementation of the proposed model by using 995 MRI image of three different subjects including 330 AD, 335 Normal Control (NC) and 330 MCI(L. Weiming et al, 2018). MCI subjects are further divided into two groups as MCI converters and MCI non-converters. Among all the used samples, 665 MRI images of 330 AD and 335 NC were used for training the system, and the remaining 330 MRI images of 166 MCI and 164 MCInc were used for testing of the system, i.e.33% of total data is used for testing purpose. The images which have been used in the experiments have the same resolution. K-fold cross-validation is used, and the value of k is 10. The data has been randomly divided into k equal sized subsamples in which one is used for training and the remaining for testing the system. The samples of the MRI image dataset is an open-source and freely available on the internet. It is downloaded from GitHub's research program named 3D-Convent-for-Alzheimer-s-Detection-master. The Convolutional Neural Network is used as a deep learning algorithm. For the experiment, a 2.2 GHz Intel Core i7 processor with 16 GB RAM is used. However, there are four categories of the images in this dataset which AD images, NC images, MCI converters images, and MCI non-converters images. All the images categories have been tested and shown in tables 3, 4, 5, and 6.

The mean and the standard deviation of the sample data of four different stages of dementia are calculated using the AVERAGE () and STDEVA () function of Microsoft Excel. The review of the literature shows that the mean and the standard deviation of the state-of-art methods for the Alzheimer's Disease, is 80.43% and 0.009064381, Normal Control is 80.64% and 0.03723681, Mild Cognitive Impairment Converters is 81.60%, and 0.010717276 and Mild Cognitive Impairment Non-converters is 80.71% and 0.008415046.

On the other hand, the proposed solution provides an improvement to the system performance which is seen clearly on the value of the mean and standard deviation for the Alzheimer's Disease is 85.98%and 0.009064381, Normal Control is 85.86%and 0.010948288, Mild Cognitive Impairment Converters is 86.26% and 0.009294246 and Mild Cognitive Impairment Non-converters is 85.75% and 0.010779239.

The formula for the standard deviation is presented in equation (5) (L. Weiming et al, 2018).

$$\sigma = \sqrt{\frac{\sum |x - \bar{X}|^2}{n}} \tag{5}$$

σ = standard deviation  
x = sample  
X̄ = mean of the samples

N = total number of samples

In the feature extraction stage, the Convolutional Neural Network extracts the features of the MRI images automatically in terms of labeled data and using the Rectified Linear Unit activation function.

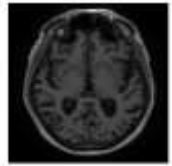

Fig. 7: Conversion to greyscale after applying the input image to Convolutional Neural Network.

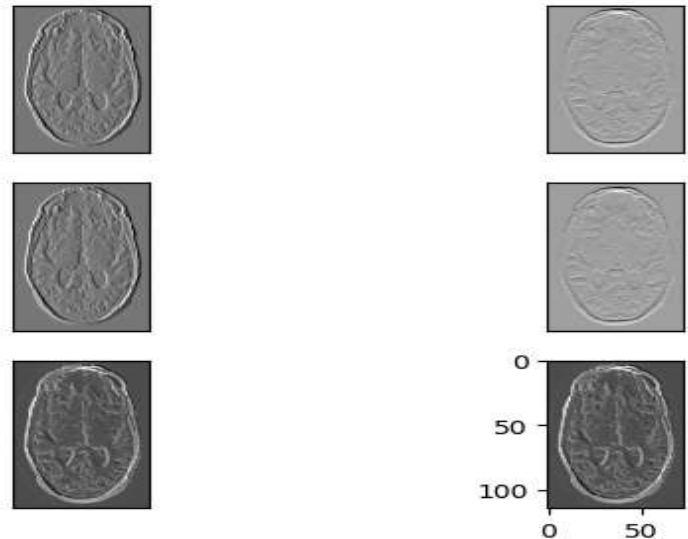

Fig. 8: Extracted the features of the input grey-scale images in Convolutional layer of the CNN

The features map output of the convolutional layer is the input of the max pooling layer of the CNN as shown in fig 9 and fig 10.

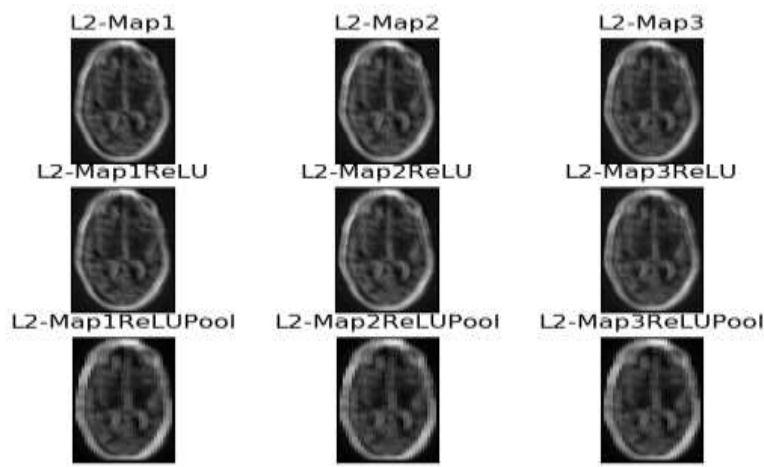

Fig. 9: Features extraction of the L2 layer of max pooling layer of CNN

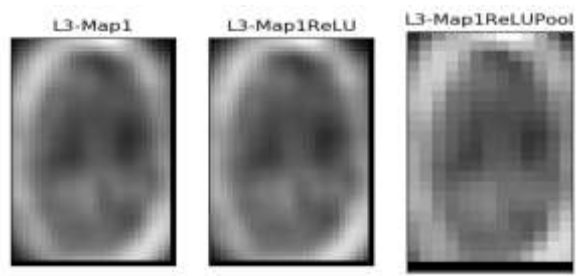

Fig. 10: Features extraction of the L3 layer of max-pooling layer of CNN

In fig 9 and fig 10, L2 and L3 are the terms to define layer two and layer three of the max pooling layer of CNN, and besides, the Map describes the features from the convolutional layer. Similarly, RELU defines the activation function for the convolutional layer, and the RELU pool defines the activation function for the max-pooling layer. The fully connected layer is used in CNN after the max-pooling layer which is then subjected to PCA and Elastic Net for features selection. Finally, the selected features are used as input to the Extreme Learning Machine to classify the MRI image. Based on the training and testing of the MRI image datasets, the sample images that have been classified as shown in tables 3, 4, 5, and 6.

To evaluate the proposed prototype and the state of the art framework the graphs and tables were used for the comparison of image sample between the proposed solution and state of the art. The obtained results after the classification of the MRI images from different groups are evaluated in table 3, 4, 5, and 6. The results are tabulated in a table according to the diverse group of MRI images. Besides, the results from the test samples are represented in terms of accuracy and processing time.

The probability score that has been labelled after the classification is used to measure the accuracy whereas the execution time for classifying the sample image after fed it to the system time is used to measure the processing time when. The test is carried out with the three different subjects of sample image data. Each test image is classified in either one of the four groups: AD subject, NC subject, MCI converters subject and MCI non-converters subject. The average of the accuracy and the average of the processing time are measured for each classified group. And then, the average of all four groups of test image data is estimated to get the overall average accuracy and average processing time of the system.

The Extreme Learning Machine is used for the classification of the image. The proposed solution has enhanced the accuracy of the classification by employing Elastic Net Regularization in the feature selection process and also has reduced the processing time by avoiding the random generation of input weight matrix in the classification process. As a result, the proposed model has improved the early prediction of Alzheimer's disease.

Table 3: Accuracy and Processing time for Alzheimer's Disease subject images (Sample image group 1)

| Sample No. | Sample group details | Original Images | State of the art | | | Proposed solution | | |
|---|---|---|---|---|---|---|---|---|
| | | | Processed sample | Accuracy (%) | Processing time (sec) | Processed sample | Accuracy (%) | Processing time (sec) |
| 1.2 | Images of Alzheimer's Disease subject | | | 79.81% | 0.353s | | 85.05% | 0.317s |
| 1.2 | | | | 79.22% | 0.347s | | 85.35% | 0.312s |
| 1.3 | | | | 81.42% | 0.344s | | 86.78% | 0.311s |
| 1.4 | | | | 80.65% | 0.309s | | 85.90% | 0.307s |
| 1.5 | | | | 81.07% | 0.355s | | 86.83% | 0.304s |

**Table 4: Accuracy and Processing time for Normal Control subject images (Sample image group 2)**

| Sample | Sample group details | Original Images | State of the art | | | Proposed solution | | |
|---|---|---|---|---|---|---|---|---|
| | | | Processed sample | Accuracy (%) | Processing time (sec) | Processed sample | Accuracy (%) | Processing time (sec) |
| 2.2 | Images of Normal Control subject | 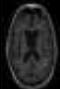 | 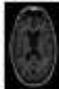 | 78.55% | 0.387s | 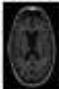 | 85.06% | 0.334s |
| 2.2 | | 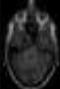 | 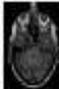 | 79.78% | 0.365s | 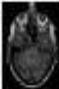 | 84.88% | 0.323s |
| 2.3 | | 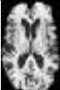 | 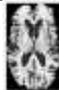 | 82.52% | 0.337s | 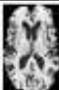 | 86.61% | 0.305s |
| 2.4 | | 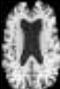 | 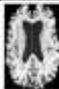 | 79.65% | 0.368s | 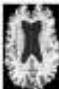 | 85.35% | 0.339s |
| 2.5 | | 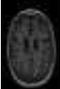 | 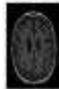 | 82.70% | 0.305s | 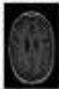 | 87.40% | 0.303s |

**Table 5: Accuracy and Processing time for Mild Cognitive Impairment Converters subject images (Sample image group 3)**

| Sample | Sample group details | Original Images | State of the art | | | Proposed solution | | |
|---|---|---|---|---|---|---|---|---|
| | | | Processed sample | Accuracy (%) | Processing time (sec) | Processed sample | Accuracy (%) | Processing time (sec) |
| 3.1 | Images of Mild Cognitive Impairment Converters subject | 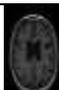 | 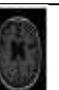 | 82.35% | 0.379s | 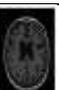 | 86.11% | 0.326s |
| 3.2 | | 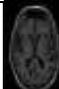 | 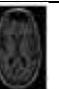 | 79.77% | 0.381s | 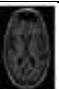 | 85.31% | 0.336s |
| 3.3 | | 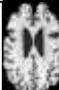 | 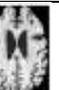 | 82.37% | 0.359s | 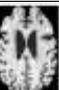 | 86.88% | 0.313s |
| 3.4 | | 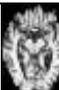 | 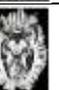 | 81.90% | 0.357s | 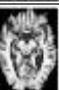 | 85.48% | 0.322s |
| 3.5 | | 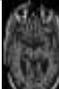 | 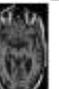 | 81.61% | 0.375s | 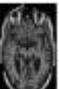 | 87.50% | 0.332s |

**Table 6: Accuracy and Processing time for Mild Cognitive Impairment Non-Converters subject images (Sample image group 4)**

| Sample No. | Sample group details | Original Images | State of the art | | | Proposed solution | | |
|---|---|---|---|---|---|---|---|---|
| | | | Processed sample | Accuracy (%) | Processing time (sec) | Processed sample | Accuracy (%) | Processing time (sec) |
| 4.2 | Images of Mild Cognitive Impairment Non-Converters subject | 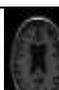 | 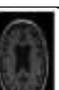 | 80.56% | 0.356s | 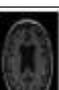 | 85.44% | 0.308s |
| 4.2 | | 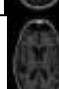 | 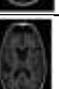 | 80.41% | 0.378s | 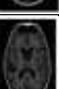 | 84.12% | 0.342s |
| 4.3 | | 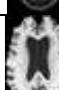 | 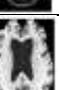 | 81.74% | 0.383s | 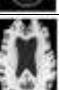 | 86.80% | 0.358s |

| | | | | | | | |
|---|---|---|---|---|---|---|---|
| 4.4 | | 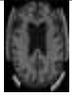 | 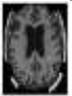 | 79.55% | 0.369s | 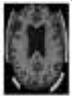 | 85.76% | 0.343s |
| 4.5 | | 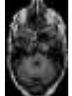 | 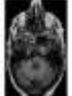 | 81.27% | 0.367s | 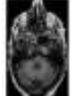 | 86.64% | 0.336s |

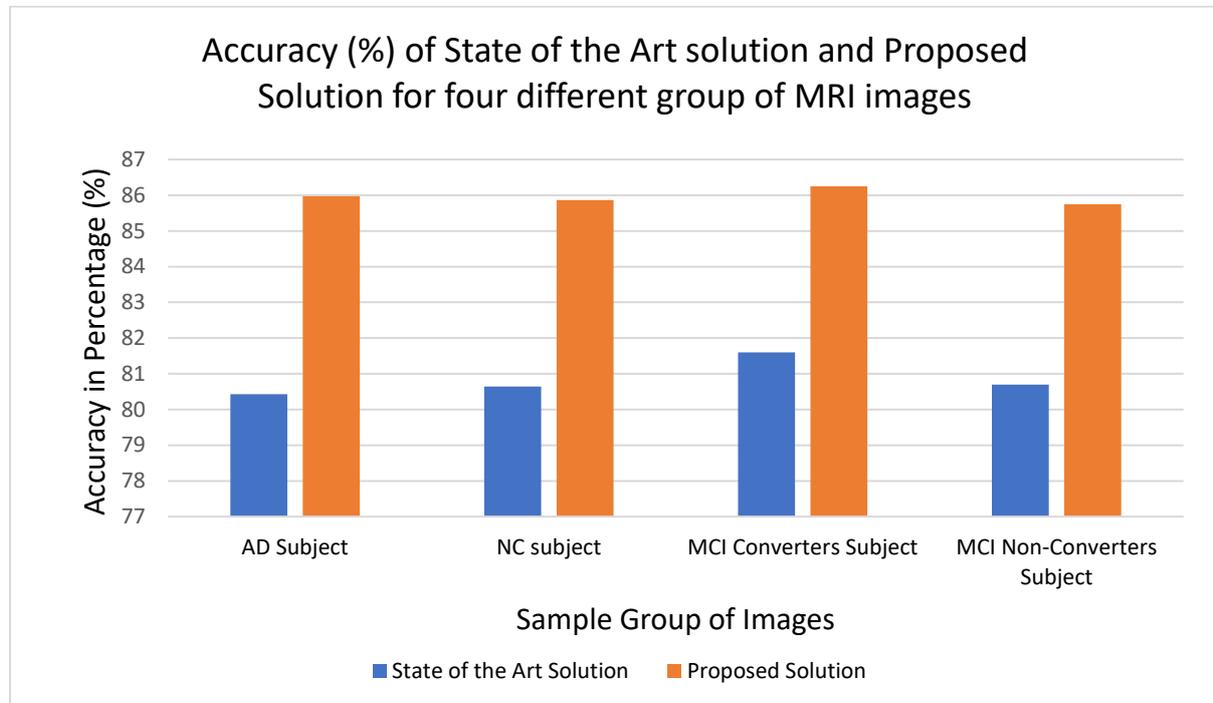

Fig. 11: The bar graph shows the classification accuracy in percentage for four different sample groups of MRI images. The blue color indicates the accuracy of the state of the art solution while the orange color indicates the accuracy of the proposed solution. (1) The first couple of bar graph shows the average accuracy for AD subject. (2) The next couple of bar graph shows the average accuracy for NC subject. (3) A third couple of bar graph shows the average accuracy for MCI Converters subject. (4) The last couple of bar graph shows the average accuracy for MCI Non-Converters subject.

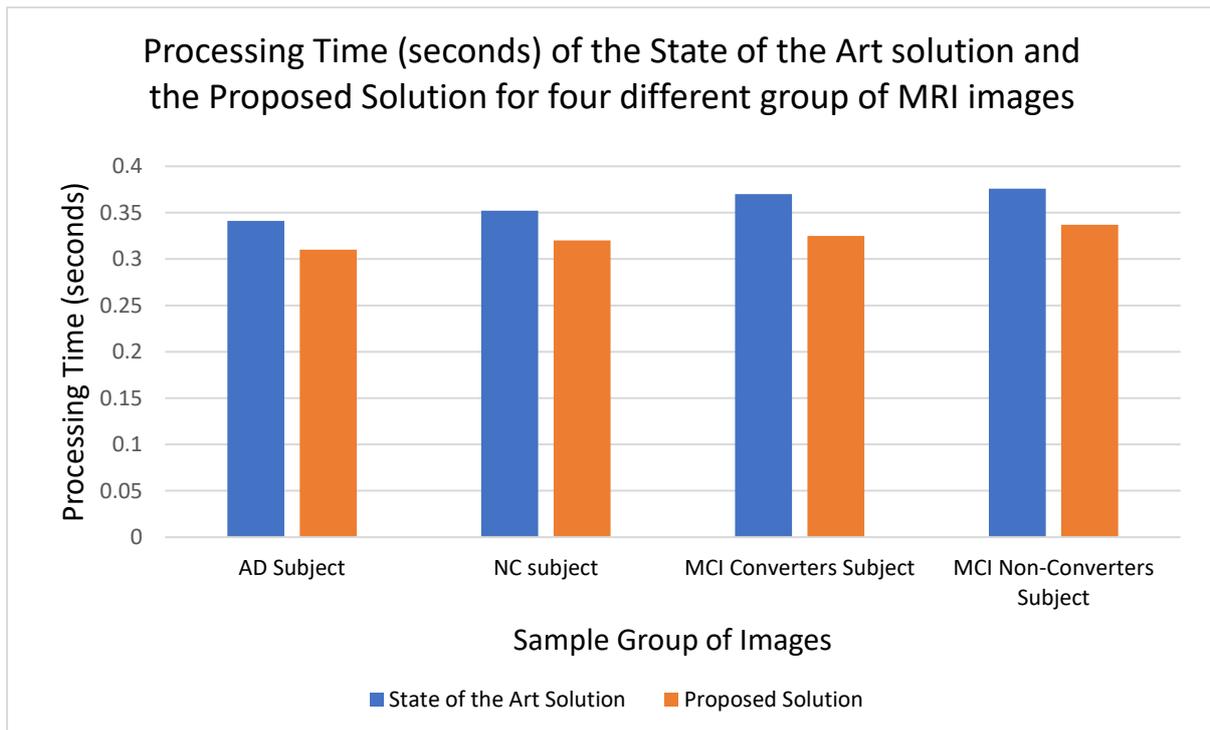

Fig. 12: The bar graph shows the processing time in seconds for four different sample groups of MRI images. The blue color indicates the processing time of the state of the art solution while the orange color indicates the processing time of the proposed solution. (1) The first couple of bar graph shows the average of the processing times for AD subject. (2) The next couple of bar graph shows the average of the processing time for NC subject. (3) The third couple of bar graph shows the average processing time for MCI Converters subject. (4) The last couple of bar graph shows the average processing time for MCI Non-Converters subject.

The results show differences in term of accuracy and processing time between the recent state of the art methods and the proposed method concerning the classification of the images. The proposed model decreases the processing time of the model by 30 to 40 milliseconds with the help of Extreme Learning Machine by avoiding the random generation of the weight matrix. Similarly, the proposed system enhances the classification by providing 86% average accuracy which is better than the state of the art solution. The proportion of true positive results is calculated to obtain the accuracy where true positive and true negative metrics are used. The probability score is used to measure the accuracy whereas execution time is used to calculate the processing time of the system. Besides, the accuracy and the processing time are calculated by using the Evaluate () method and the now () method of the Keras Python package. The now () method produces the execution time through calculate the subtraction between the state time and the end time of the process. Processing time is quantified by calculating the time duration of running the proposed solution algorithm and state of the art algorithm. As shown above, the accuracy metric is measured in percentage while the processing time is measured in second.

Apart from that, the Elastic Net Regularization is used in the feature selection stage after applying the Convolutional Neural Network-based feature which has improved the performance of the proposed system. The proposed system uses Python programming language for the implementation which has improved the classification accuracy and minimizes the processing time of the model. The use of Elastic Net Regularization forms a group of correlated variables instead of taking only one variable and includes all the relevant independent variables in the feature extraction stage, and this will lead to an increase in the accuracy of the proposed system. Besides, the use of the Extreme Learning Machine for the classification has minimized the processing time by avoiding the random generation of the weight matrix. In conclusion, the Convolutional Neural Network combined with the Elastic Net Regularization and Extreme Learning Machine has enhanced the MRI image classification accuracy by 5% higher than the state of the art method and processing time by 30s ~ 40s in overall.

Last but not least, various techniques and algorithms have been implemented for the classification of the MRI images. However, continuous refinement of the techniques has been carried out to improve the accuracy and processing time. The limitation of the state of the art has been solved in this research with improved accuracy of 86% against the current accuracy of 81%. The system also reduces the processing time to 323 milliseconds against the 359 milliseconds. The improved results have been achieved due to the implementation of the Elastic Net Regularization in the feature selection stage to include all the important and relevant independent variables as well as correlated variables of the extracted features. To conclude, the proposed system has better accuracy and reduced the processing time even when it applied to different image groups such as images of AD subject, images of NC subject, images of MCI Converters subject, and images of MCI Non-Converters subject.

## 5. Conclusion and Future Work

Accurate classification of MRI images is essential to predict MCI to AD conversion accurately. Deep learning has been successfully implemented to classify and predict dementia disease. However, there are some limitations that effect in accuracy and processing time of the system. The purpose of this paper is to improve the accuracy and to reduce the processing time of MCI converters/non-converters classification by using deep learning techniques. The Elastic Net Regularization has been developed by combining the Lasso regression and modified Ridge regression which is adapted from the second-best solution (M. Liu et al, 2018). It adds a quadratic part to the penalty which is obtained after modifying the Ridge Regression. It forms a group of the correlated variable instead of selecting only one variable so that important features are not neglected or missed during feature selection. It also maintains the consistency in feature selection by including all the relevant independent variables. Hence, it enhances the classification accuracy by 5% on average and reduced the processing time by 30 ~ 40 seconds on average. See Table 7.  In future work, a large amount of MRI image datasets can be used for the training process of CNN. More features can be extracted from CNN to train the model effectively which could further enhance the performance of the proposed system. More feature selection methods will be implemented and compared with the proposed system.

**Table 7: Comparison table of State-of-art and Proposed solutions**

| | Proposed Solution | State of the Art Solution (L. Weiming et al, 2018) |
|---|---|---|
| **Name of the solution** | Elastic Net Regularization | Lasso Regression |
| **Accuracy** | Improved the accuracy in terms of feature selection. Provides an accuracy improvement in classification of 86%. | Improved Accuracy in terms of MCI to AD conversion's prediction. Provides an accuracy of 81.4% and Area under the curve of 87.7%. |
| **Processing Time** | The decrease in processing time from 359ms to 323ms on average. | Provides a processing time of 359ms on average. |
| **Proposed equation** | $EL(W) = L(W) + MR(W)$ <br> Where, <br> EL(W) is the Elastic Net Regularization, <br> L(W) is the Lasso Regression and <br> MR(W) is the Modified Ridge Regression | $L(W) = \min_{\propto} 0.5\|y - D \propto\|_2^2 + \lambda\| \propto \|^1$ <br> Where, <br> $y \in R^{1 \times N}$ is the vector consisting of $N$ labels of training samples, <br> $D \in R^{N \times M}$ is the feature matrix of $N$ training samples comprising of $M$ features, <br> $\lambda$, is the penalty coefficient <br> $\propto \in R^{1 \times M}$ is the sparse target coefficient |
| **Contribution 1** | Elastic Net Regularization adds a quadratic part to the penalty to include all the relevant independent variables. | The state of the art method does not include all the relevant independent variables. |
| **Contribution 2** | It forms a group if there is a correlated variable to protect the valuable information from being removed. | The state of the art selects only one variable from a group of highly correlated variables and ignores others. |

**Compliance with Ethical Standards:**

Funding: No Funding has used in this work.
Conflict of Interest: No conflict of interest

# Appendix

**Table 8: Abbreviation for term used in Analysis Table**

| | |
|---|---|
| **DVIC** | Depth Visual Information Construction |
| **ADNI** | Alzheimer's Disease Neuroimaging Initiative |
| **DPABI** | Data Processing & Analysis of Brain Imaging |
| **PCA** | Principal Component Analysis |
| **CNN** | Convolutional Neural Network |
| **EI&ES** | Electrode Implantation & Electrical Simulation |
| **DFL** | Deep Feature Learning |
| **SL&LR** | Semi-Supervised Learning & Logistic Regression |
| **AHC** | Agglomerative Hierarchical Clustering |
| **LASSO** | Least Absolute Shrinkage and Selection Operator |

Kshitiz Shrestha, Omar Hisham Alsadoon, Abeer Alsadoon (Corresponding author), P.W.C. Prasad, Rasha S. Ali, Tarik A. Rashid, Oday D. Jerew. (2021). A Novel Solution of an Elastic Net Regularization for Dementia Knowledge Discovery using Deep Learning. Journal of Experimental & Theoretical Artificial Intelligence.